\documentclass[lettersize,journal]{IEEEtran}
\usepackage{amsmath,amsfonts}
\usepackage{algorithmic}
\usepackage{algorithm}
\usepackage{array}
\usepackage[caption=false,font=normalsize,labelfont=sf,textfont=sf]{subfig}
\usepackage{textcomp}
\usepackage{stfloats}
\usepackage{url}
\usepackage{verbatim}
\usepackage{graphicx}
\usepackage{cite}
\usepackage{color}
\usepackage{listings}

\definecolor{jsonKeyColor}{rgb}{0, 0, 0.5}
\definecolor{jsonValueColor}{rgb}{0, 0.5, 0} 
\definecolor{jsonBraces}{rgb}{0.5, 0, 0} 

\lstdefinestyle{jsonStyle}{
  showstringspaces=false,
  basicstyle=\ttfamily\small,
  stringstyle=\color{jsonValueColor},
  keywordstyle=\color{jsonKeyColor},
  commentstyle=\color{jsonBraces},
  upquote=true,
  morestring=[b]",
  morecomment=[s]{\{}{\}},
  morecomment=[s]{[}{]},
  morekeywords={null},
  literate=
    *{0}{{{\color{jsonValueColor}0}}}{1}
     {1}{{{\color{jsonValueColor}1}}}{1}
     {2}{{{\color{jsonValueColor}2}}}{1}
     {3}{{{\color{jsonValueColor}3}}}{1}
     {4}{{{\color{jsonValueColor}4}}}{1}
     {5}{{{\color{jsonValueColor}5}}}{1}
     {6}{{{\color{jsonValueColor}6}}}{1}
     {7}{{{\color{jsonValueColor}7}}}{1}
     {8}{{{\color{jsonValueColor}8}}}{1}
     {9}{{{\color{jsonValueColor}9}}}{1}
     {:}{{{\color{jsonBraces}{:}}}}{1}
     {,}{{{\color{jsonBraces}{,}}}}{1}
     {\{}{{{\color{jsonBraces}{\{}}}}{1}
     {\}}{{{\color{jsonBraces}{\}}}}}{1}
     {[}{{{\color{jsonBraces}{[}}}}{1}
     {]}{{{\color{jsonBraces}{]}}}}{1},
}

\begin{document}

\title{Zero-Shot Reasoning: Personalized Content Generation Without the Cold Start Problem}

\author{
    \IEEEauthorblockN{Davor Hafnar\textsuperscript{1} and Jure Demšar\textsuperscript{1,2}}\\
    \IEEEauthorblockA{\textsuperscript{1}Faculty of Computer and Information Science, University of Ljubljana, Ljubljana, Slovenia}\\
    \IEEEauthorblockA{\textsuperscript{2}Department of Psychology, Faculty of Arts, University of Ljubljana, Ljubljana, Slovenia}
}

\maketitle

\begin{abstract}
Procedural content generation uses algorithmic techniques to create large amounts of new content for games at much lower production costs. To improve its quality, in newer approaches, procedural content generation utilizes machine learning. However, these methods usually require expensive collection of large amounts of data, as well as the development and training of fairly complex learning models, which can be both extremely time-consuming and expensive. The core of our research is to explore whether we can lower the barrier to the use of personalized procedural content generation through a more practical and generalizable approach with large language models. Matching game content to player preferences benefits both players, by enhancing enjoyment, and developers, who rely on player satisfaction for monetization. Therefore, this paper introduces a new method for personalization by using large language models to suggest levels based on ongoing gameplay data from each player. We compared the levels generated using our approach with levels generated with more traditional procedural generation techniques. Our easily reproducible method has proven viable in a production setting and outperformed levels generated by traditional methods in two aspects – the player's rating of levels and the probability that a player will not quit the game mid-level.
\end{abstract}


\section{Introduction}
\IEEEPARstart{M}{obile} game market is highly saturated and simply creating a good game is no longer enough to succeed. Besides a good game, you usually also need a sizeable advertising investment. Because of this, the cost per install can be high and retaining players is very important. We can keep the player in the game by providing new and engaging content, which can again be expensive, as its production requires a lot of time from developers, artists and game designers. A popular way to address this is through procedural content generation (PCG).

PCG uses algorithmic techniques to create content for games. It is employed to increase replay value, reduce production costs and effort, or save storage space. Apart from accounting for difficulty, it is usually not personalized – content is not generated to match the preferences of a specific player. In other words, procedurally generated content is the same for all players, regardless of their play style.

Some interesting work has already been done in the area of game personalization, such as the proposal of Play Data Profiling (PDP) framework \cite{rajanen2017personalized}, which included psychophysiological measurements and eye tracking for the purpose of adjusting the game to the given player's current state. There has also been some industry-driven research, mostly done by game studios \cite{xue2017dynamic}, whose data science teams aim to improve the key performance indicators (KPIs) of their games. Only a limited amount of their research appears to be published and even in those cases, implementation details are very scarce.

Our idea is to leverage the latest developments in machine learning (ML) to develop a new type of PCG framework. Most recent large language models (LLMs) have introduced the ability for few-shot learning and zero-shot reasoning, meaning that they can be used for solving tasks they were not specifically trained for. This way they avoid the cold start problem, which presents a significant barrier to the integration of ML in recommendation systems. Cold start problem refers to the challenge of providing useful recommendations to players without first accumulating a sufficient amount of data \cite{Panteli2023coldstart}. One recent approach to accumulating a substantial volume of data, to ensure that players receive high-quality recommendations from the start, is through the generation of synthetic data, often employing deep reinforcement learning techniques \cite{shin2020playtesting, gudmundsson2018human}. However, even after addressing the cold start problem, there remains the necessity to train task-specific models, which can be a costly endeavour. This training process does not generalize and must be repeated for each game or even for major updates within the same game.

These problems are not present in state-of-the-art LLMs, which have so far mostly been used to generate human-like text \cite{dale2021gpt}. In the context of PCG, even though they are a relatively new technique, LLMs have been already successfully used to generate texts for quest-based games \cite{vartinen2022generating} and, with human supervision, entire levels \cite{todd2023level}. As LLMs get more powerful and tuned to flawlessly provide computer-readable output, such as the JSON format \cite{Eleti2023FunctionCalling}, they can be, as we validated in our research, used to generate personalized level parameters.

The purpose of our study is threefold:
\begin{enumerate}
    \item To validate that modern LLMs (for example GPT-4) can be used in production for personalized PCG with no additional training and without human-in-the-loop with high reliability.
    \item To verify if through zero-shot reasoning capable LLMs, ML can be used in personalized PCG without a cold start problem, which would lower the barrier to using personalized PCG in games.
    \item To compare our approach with traditional PCG, by analyzing play data in order to check whether our approach can outperform traditional PCG when it comes to players completing the levels they started.
\end{enumerate}

\section{Related work}
PCG has been discussed in the literature for decades \cite{Hendrikx2013},\cite{Perlin1985}. One of its subareas is called personalized PCG, which is about generating content for individual players based on their skills, style and preferences \cite{liu2021deep}. In recent years, approaches that incorporate ML \cite{Summerville2018}, deep learning \cite{liu2021deep}, and more recently LLMs \cite{todd2023level, sudhakaran2023prompt} have started to gain popularity. Our research is positioned in the area of LLM-based personalized PCG. 

\subsection{Personalized PCG}

Personalized PCG, sometimes also referred to as adaptive or player-driven PCG \cite{togelius2011whatispcg}, is a less explored area of PCG \cite{liu2021deep}. A possible design for such games was described by Rajanen and Rajanen \cite{rajanen2017personalized}. Their idea is that a gaming system should be collecting real-time play data, which can then be used for player profiling even after the system has been developed. They introduce a PDP model which proposes that gamification elements are adapted based on the data derived from the interaction and the personal data of the player. A periodic reassessment of the player may determine that a player is moved from one profile cluster to another. Our research puts this idea into practice. 

A more limited personalized approach aimed at player retention was utilized by Milošević et al. \cite{milovsevic2017early}. They focused on retaining players in a mobile game by utilizing early churn prediction and personalized player targeting. They first predicted which players were likely to churn and sent each one of them a personalized notification. They determined that such a personalized approach can retain players who would have otherwise left the game. Researchers at Electronic Arts \cite{xue2017dynamic} tackled dynamic difficulty adjustment as an optimization problem. The Match 3 game generated random levels with varying difficulty. The goal of ML-based optimization in their case was to maximize player engagement over the entire game. Their solution increased core engagement metrics such as rounds played and gameplay duration, however, it only focused on adjusting the difficulty, while our research aims to facilitate broader personalization.

\subsection{Large Language Models}

In literature, research on the use of LLMs often leverages an OpenAI GPT-2 model, which is open-source and allows fine-tuning \cite{radford2019gpt2}. Van Stegeren and Myśliwiec\cite{van2021fine} investigated the usability of fine-tuned GPT-2 for dialogue line generation for a quest giver in a role-playing game. They tested the quality of generated quests with human judges. Even though GPT-2-generated quests on average performed worse than human-generated ones, authors found the method a viable option for generating text in games. A more extensive study in the field of quest generation was performed by V\"artinen et al. \cite{vartinen2022generating} who explored the possibility of quest generation using GPT-2 and GPT-3. They fine-tuned GPT-2 with additional data and asked players for feedback. They concluded that GPT-2 was not yet up to the task, but future models like GPT-3 are likely to provide the ability to generate high-quality content autonomously. 

The use of LLMs to generate levels for the game Sokoban was tested by Todd et al. \cite{todd2023level}. They devised a scoring model to determine the level's novelty, playability, diversity and score. They concluded that, despite it being a very different domain from natural language, the use of LLMs for game-level generation shows promise. Similarly, the study by Sudhakaran et al. \cite{sudhakaran2023prompt} presents MarioGPT, a GPT-2 model proficient in procedural content generation. The study exhibits MarioGPT's capability to generate diverse gaming environments. Generated levels were playable approximately 88\% of the time. Prompt engineering as a technique for generating game levels was showcased in the ChatGPT4PCG Competition\cite{taveekitworachai2023chatgpt4pcg}. In this event, participants competed to create effective LLM prompts to generate levels for a game similar to Angry Birds. While studies on LLMs still mostly dealt with the feasibility of LLMs for PCG, the goal of our research is to test the feasibility of a reproducible, production-level framework.

LLMs have in recent years revolutionized and dominated the field of natural language processing (NLP) \cite{chernyavskiy2021nlp}. They have been used for tasks such as dialogue systems, text summarization and machine translation \cite{openai2023gpt4}. In NLP, text generation is typically approached by assigning the probability of the next token based on the preceding tokens. Statistical approach can be described as \[ c_i \sim p(c_i \mid c_1, \ldots, c_{i-1}; \theta), \] where $c_i$ denotes the $i^\text{th}$ token in the text sequence and $\theta$ represents the parameters of the sampling distribution. The task is to maximize the probabilities of all tokens in the training data \cite{hamalainen2023llm, vartinen2022generating}. A token can be a word, a part of the word or a letter. For example, given the words ``Grand Theft'', a model pre-trained on text found on the internet is likely to assign the next word ``Auto'', as this word is most likely to follow two preceding words.

The development of Generative Pre-Trained Transformer (GPT) by OpenAI, starting with the original GPT \cite{radford2018gpt}, followed by GPT-2 \cite{radford2019gpt2}, GPT-3 \cite{Brown2020gpt3} and most recent GPT-4\cite{openai2023gpt4} introduced models that can generate coherent and contextually relevant text across a wide range of topics. These models are based on the transformer architecture, which makes training more parallelizable and significantly faster than NLP architectures before them \cite{vaswani2017attention}. 

\subsection{Zero-shot reasoning}

The pre-training of LLMs involves learning to predict the next word in a sentence by consuming vast amounts of text data. This enables the model to capture a wide range of linguistic patterns and world knowledge. This pre-training can be followed by fine-tuning for specific tasks. Even without fine-tuning, GPT-3's and GPT-4's size allows for \textit{few-shot learning} or \textit{zero-shot reasoning}, where the model can perform tasks with little to no task-specific training data \cite{Brown2020gpt3}. 

Historically, \textit{zero-shot learning} has been used to refer to classifying instances without requiring training samples of the target classes \cite{lampert2009unseen}, translating between unseen language pairs \cite{johnson-etal-2017-googles} and modelling on unseen languages\cite{lauscher-etal-2020-zero}. The term is recently used to describe how models generalize to unseen tasks, an emerging ability of language models \cite{wei2022finetuned}. In the paper introducing GPT-3, researchers have shown that the 175 billion parameter language model has shown performance almost matching state-of-the-art fine-tuned systems on many NLP tasks in zero-shot, one-shot and few-shot settings \cite{Brown2020gpt3}. While the term \textit{zero-shot learning} is used also in the context of LLMs \cite{wei2022finetuned}, we prefer to use the term \textit{zero-shot reasoning} \cite{kojima2022zsreason, wan-etal-2023zsreason}.

For our experiment, we used GPT-4, a state-of-the-art successor to GPT-3. While it is a proprietary model whose inner workings are not published, it outperforms the predecessor on most tasks \cite{openai2023gpt4}. In the context of PCG, the zero-shot performance of capable LLM models can be used to generate level parameters with minimal examples or guidance. This is particularly useful for indie developers or small studios that may not have extensive datasets to train more traditional ML models.

\section{Approach and basic components}

For our research, we have developed and published a basic Match 3 mobile game. The game was developed in C\# programming language using the Unity game engine \cite{haas2014unity} and interacted with a Google Cloud micro services-based back end. Products from the Google Cloud such as Cloud Functions and BigQuery were used for data collection. The machine learning model used for personalization was an LLM provided by OpenAI called GPT-4 \cite{openai2023gpt4}, with which we interacted using a REST API under commercial terms. Furthermore, GPT-4's function calling capability allows us to receive its outputs in the JSON format, which is much easier to use for our purposes.

The game was published on the Google Play Store and players were acquired using Google Ads as is traditional for any commercially available game. This approach enabled us to obtain a representative sample of players, which would have been more challenging to achieve had we recruited participants ourselves. 

\subsection{The match 3 game}

\begin{figure}[htbp] 
 \centering 
 \includegraphics[width=\columnwidth]{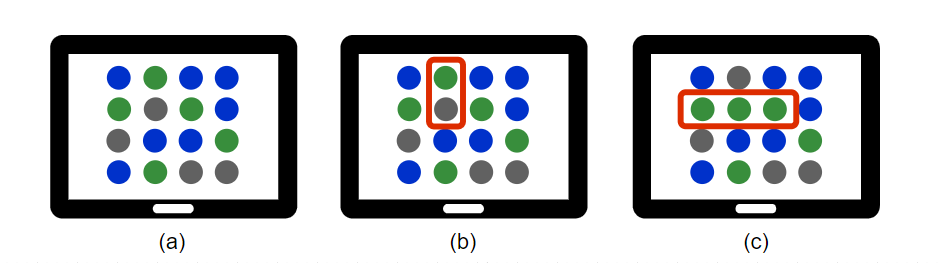} 
\caption{\textbf{An illustration of game mechanics in our Match 3 puzzle game.} (a) A board is filled with pieces of different colours. (b) The player's objective is to find and match three board elements of the same colour. He can do that by swapping neighbouring pieces. (c) Once three or more elements of the same colour form a line, the player is awarded a certain amount of points, while the matching elements are removed from the board and replaced by new elements.} \label{fig1}
 \label{fig:match3}
\end{figure}

Our game uses the Match 3 mechanics, which has already been proven as suitable for such research by Xue et al. \cite{xue2017dynamic}. The basic gameplay is visualized in Figure~\ref{fig1}. Before starting a level, a player is shown a pop-up with objectives and limitations for the level. An example of an objective is a score that the player has to reach, combined with the number of elements of a specific colour he needs to connect and consequently remove from the board. Additionally, the game includes boosters, which are special items that can be used to clear more pieces or overcome challenging obstacles. The player is constrained by the number of moves: he needs to use the available moves strategically to clear the assigned amount of pieces in a certain colour as well as collect enough points to complete the level.

The popularity of Unity, a game development engine, as well as the popularity of the Match 3 genre, helped with development as multiple useful templates are readily available. Our game was based on one such template \cite{lin2020match3}.

\subsection{Data collection}
The gameplay data describes how a player interacts with the game. They are acquired through gameplay, the more the player plays the more of his gameplay data we have.

The mobile game collected gameplay data and sent it to our back-end system periodically. The most important data for level generation was collected upon the completion of each level. At that time, the player was asked to rate the level they just completed on a scale from 1 to 5. After they rated the level, the data was sent to the backend. The data collected included the number of the level completed, remaining moves, number of failed moves, overall clicks on the board, number of boosters used and how the player rated the level. All parameters, along with their description, are listed in Table~\ref{table:game_level_data}. Additionally, parameters of the level, such as the number of different pieces, board width and height, collection goals for individual pieces, along with the score required to pass and the allowed number of moves, were also collected.

For our analyses, we also collected data about players starting and completing a level, allowing us to calculate the completion rate. Additionally, to be able to track their progress, we wanted to know at which screen players are and send this data in real-time. This data helped us determine potential bottlenecks in the game. Our design focused on gameplay data, so we did not collect or use any demographic or personal data from the players.

\begin{table}[!htbp]
\renewcommand{\arraystretch}{1.3} 
\caption{Player Data Metrics and Level Parameters}
\label{table:game_level_data}
\centering
\begin{tabular}{p{2.5cm} p{5.5cm}}
\hline
\multicolumn{2}{c}{\textbf{Player Data Metrics}} \\ \hline
\textbf{Field Name} & \textbf{Description} \\ \hline
level\_in\_row & Level number for the user. \\
score & Score earned in this level attempt. \\
moves\_left & Number of moves remaining. \\
num\_failed\_moves & Number of illegal moves where pieces were moved back as no match was made. \\
num\_clicks\_on\_board & Total number of clicks on the board. \\
num\_boosters\_used & Number of times the user used a booster. \\
user\_rating & User's rating of the level. \\ \hline
\multicolumn{2}{c}{\textbf{Level Parameters}} \\ \hline
\textbf{Field Name} & \textbf{Description} \\ \hline
num\_different\_pieces & Number of different types of pieces on the board. \\
score\_goal & The score goal that needs to be achieved to pass the level. \\
board\_width & The width of the game board. \\
board\_height & The height of the game board. \\
num\_moves & The number of moves allowed to achieve the score goal. \\
collection\_goals & The collection goals in terms of the number of specific pieces that need to be collected. \\ \hline
\end{tabular}
\end{table}

\subsection{Prompting the LLM}\label{chap:promptingllm}

We used the gameplay data to generate plaintext descriptions and sent them to the GPT-4 LLM. The data for at most last five levels was sent each time the player completed a level. The gameplay data, along with hardcoded instructions, were used to prompt the model to generate the next three levels and return them as JSON-formatted level parameters. The response itself was JSON formatted as the prompt dictated the following steps:

\texttt{
Your task is to:
\begin{enumerate}
\item Consider the data on the player.
\item Consider parameters of the levels the player already completed.
\item Determine what type of player we are dealing with based on a list of player types. Mostly consider the level of skill, fun vs. complex, puzzly vs arcade.
\item Suggest the next 3 levels for this player based on the type of gamer and the list of level completion parameters.
\item Explain your reasoning for the type of gamer and next 3 levels.
\end{enumerate}}

We also provided more precise instructions about how to select the parameters:

\texttt{
\begin{itemize}
\item level\_number: A number of the current level.
\item num\_different\_pieces: More different pieces, harder the game. Valid range (3, 5). 
\item score\_goal: The score a user must reach before completing the level. The score should be divisable by 3. Valid range (700, 2000). 
\item num\_moves - Amount of moves a user has to complete the game. Harder levels need more moves: consider collection\_goals. Usually number is (20, 30). 
\item board\_width - How wide the board is. Wider is harder. Valid range (4, 6). 
\item board\_height - Height of the board. Higher is harder. Should be very similar to board-width. Valid range (4, 6). 
\item collection\_goals - To finish the level, you need to collect a certain number of pieces with a specific color. \\
\end{itemize}}

The game involves (2, 4) unique colors, each represented by a number of pieces between (5, 15).
For instance, if the game requires two different colors, you may need to collect 10 pieces of one color and 20 pieces of another, represented as [10, 20].

To ensure valid levels and consistent results, we utilized the \textit{function calling} capability of GPT-4, which is specifically designed to provide results in a consistent JSON format \cite{OpenAIFunctionCalling}. We input similar descriptions and limitations into the function calling feature to maintain uniformity in our approach.

The first two steps instructed the model to incorporate data from the player's device regarding completed game levels. In the third step, the model was tasked to categorically assign players to predefined types based on the player's skill and playstyle (\textit{not so skilled player}, \textit{casual player}, \textit{great player}).

In the fourth step, we asked the model to propose the next three levels for the assigned player type within the predetermined parameters. We did this after each level, appending the gameplay descriptions of at most five last completed levels. In addition, for analytical purposes, we asked the model to provide the reasoning for the player type selection and level parameters. We logged any additional data returned from the model for further analysis and review. We decided to generate three levels and not just one to make sure there were always some levels ready for the player in case of latency or backend technical difficulties.
An example of how a prompt with two levels was sent to the LLM:

\texttt{For level 3, the user scored 1460 where 900 was the minimum to pass. They had 18 moves left out of 25. They made 3 failed moves. They made 59 clicks on the board. They used 0 boosters. The player rated the level as 1 out of 5. The level contained 3 different pieces. Board width x height was 4 x 4. }

\texttt{For level 4, the user scored 1860 where 800 was the minimum to pass. They had 18 moves left out of 25. They made 0 failed moves. They made 67 clicks on the board. They used 0 boosters. The player rated the level as 5 out of 5. The level contained 3 different pieces. Board width x height was 4 x 4."}

In this case, the player was assigned to be a \textit{casual player} and the following explanation was given:

\texttt{"The player consistently completes levels with a significant number of moves left, indicating a good understanding of the game mechanics. However, the player's high number of clicks and occasional failed moves suggest they are not overly concerned with efficiency or perfect play. The player's ratings indicate they prefer levels that are not too challenging, as they rated the level they performed best on (level 4) the highest."}

To maximize the consistency of the model outputs in relation to the inputs, we set the temperature parameter of the GPT-4 API to the lowest possible value, $0$. Lower temperatures lead to more uniform outputs, whereas higher temperatures produce results that are more varied and imaginative \cite{OpenAI2023TemperatureParameter}. The returned values remained consistent for the same input, but we did notice that the target score needed to complete the level varied by up to 100 points. The target score had a valid range from 700 to 2000 points.

We decided to construct human-readable prompts, as this did not produce much overhead, while it simplified the debugging and analysis. As we were able to provide the model with a concise summary of the gameplay for the last five levels, we decided against using the GPT-4 model in a chatbot-like capacity where we would keep and append the entirety of all past interactions to each new call to the model.

\subsection{The Procedural content generation module}

We designed our game to generate levels based on selected key parameters. Those input parameters for the generation of a level for a specific player were provided by the LLM. For example, if the game was played by a lower-skilled player, the model might recommend parameters such as a smaller size grid, fewer pieces and more available moves. Apart from delivering a personalized experience, a PCG model helps keep the game fresh as levels keep changing.

\begin{figure}
  \includegraphics[trim=18 18 18 18,clip,width=\columnwidth]{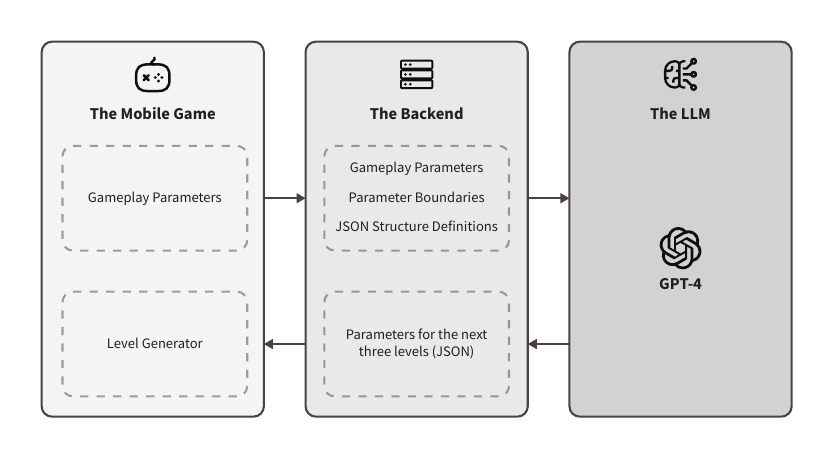} 
\caption{\textbf{Personalized PCG Pipeline Diagram.} Once the player completes and rates a level, the data is sent to the backend server. There, the prompt is constructed and level parameter boundaries are described. The strict JSON is enforced by utilizing the function calling functionality of GPT-4 LLM API. The data is then sent to the LLM, and three levels are returned as JSON. The mobile game then generates the levels based on suggested parameters. Ideally, the player plays just the first of three levels and a new batch of levels is generated after he completes that one, but three are returned to account for possible latency as level parameters are not generated on the device.} \label{pipeline}
\end{figure}

We generated the next three levels on the backend every time the player completed and rated a level. For personalized PCG, we considered the gameplay data of the last five completed levels. Levels were generated within given parameters and were for our experiment either generated using the LLM or a uniform random algorithm. Ranges for parameters were the same for both groups. Parameters for the levels were the number of different pieces in the level, board width and height, overall score goal, individual goals per piece and the number of moves allowed to complete the level. The colours of the board pieces and the colours of the pieces the player had to collect were randomly assigned each time the player started the level. Based on these parameters, our game was able to generate a level and serve it to the player. The diagram of the described pipeline is presented in Figure~\ref{pipeline}.

\subsection{Personalized gameplay}

One of the primary features of our framework is its ability to gather real-time gameplay data, which it uses to continuously personalize the experience as the player progresses through the game. Although all players begin at a similar starting point, their gameplay experiences will diverge increasingly as they play more. Our framework is designed to offer a game that presents an appropriate challenge and style for both a young child and an adult familiar with the genre and seeking a challenge. By ensuring the game is tailored to the player, we also maximize the likelihood of retaining the player.

\section{Experiment Design}

There is little incentive to implement personalized PCG if the produced personalized content does not bring any measurable benefits. To validate our approach, we used a method known as statistical testing (also known as AB testing), where players are assigned to groups that are served different versions of the game. In our case, the difference was in how we generated levels. The superior method is determined based on how well it is capable of retaining the players, meaning that players do not abandon the levels they started. We also asked the players to rate each completed level. 

Multivariate testing helped us determine if serving personalized levels outperforms randomly served levels. We distributed our game to players who were split into two groups as follows:

\begin{enumerate}
  \item players were served traditionally generated levels (procedural content generation with randomized parameters within certain thresholds), 
  \item players were served personalised levels based on parameters generated by the LLM.
\end{enumerate}

The group the player belongs to was determined the first time he opened the game. Upper and lower bounds for specific parameters were the same for both groups: the game included 3 to 5 different pieces and 2 to 4 goals, with goal scores ranging from 700 to 2000 points. Players had 20 to 30 moves to play on a board sized between 4x4 and 6x6 units. For collection goals, gathering 5 to 15 of each piece was necessary.

\subsection{Serving initial levels}

As the model generates subsequent levels based on the gameplay data from at least one completed level, a different approach was needed for the first level a player gets. If the player was assigned personalised levels, they were generated using a similar prompt as before, just instead of multiple steps as described in Chapter \ref{chap:promptingllm}, we used a simpler prompt: 

\texttt{"Your task is to suggest 3 levels of a game to a player that is completely new to it and starts with level 1."}

For players assigned random levels, level generation was the same for the first level and for levels that followed. For both groups, the levels were regenerated each time they were downloaded by a player's mobile device.

\subsection{Participants}

As we wanted the population of our players to be unbiased and reflective of the real world, we used the same approach as games traditionally used when published. Our game is publicly available through the Google Play platform. To acquire players, we used paid advertising through the Google Ads system. We invested on average 3€ per day for 26 days. Through this process, we onboarded 102 unique players who completed at least one level. Of those 102, 59 were assigned to the group using LLM-based PCG, and 43 to the group using traditional PCG. Although the assignment to each group was equally likely, our analysis shows that the group with LLM-based PCG was more likely to complete the first level, hence the observed difference. Together, they played 928 levels and completed 422 of them.

\subsection{Statistical analysis}

We used Bayesian statistics to analyse the results. All analyses were conducted using Stan -- a state-of-the-art platform for executing modern Bayesian statistical analyses \cite{StanDevelopmentTeam}.

To analyse if players were more likely to complete levels that were generated with LLM-empowered procedural content generation in comparison to traditional approaches, we facilitated a simple Bernoulli model:

\begin{equation}
    y \; | \; \theta \sim \text{bernoulli}(\theta).
\end{equation}

Values 1 in the input data vector ($y$) denote successful completions of a level, while values 0 denote unsuccessful completions, either because of running out of moves or giving up. Stan’s default non-informative prior was used for the $\theta$ parameter. We fit the above model to both groups and then compared posterior parameters to determine whether there is a difference between the completion rate and to determine the certainty of our claims.

To analyse if player ratings between the two groups differ, we used an ordered generalized linear model (GLM) \cite{bscourse2023}:

\begin{equation}
    y \; | \; \beta, x, c \sim \text{ordered\_logistic}(x \beta, c),
\end{equation}

\begin{equation}
    \beta \sim \text{Cauchy}(0, 2.5).
\end{equation}

where $y$ are player ratings, $x$ is the independent variable (the group) in our case, $c$ are the model's intercepts (or cutpoints) and $\beta$ is the regression coefficient. We put a default weakly informative Cauchy prior on the beta coefficient \cite{Gelman2008Weakly}. We use the $\beta$ parameter to estimate whether ratings in one group were larger than the other as well as the certainty of our claim.

To distinguish reported Bayesian probabilities from frequentist p-values we denote them with a capital P. Unlike p-values, the reported probabilities directly describe the probability by which we can claim that our hypotheses are true or not. The probability that the opposite of our claim is true can be calculated as $1 - P$. We used Monte Carlo Standard Error (MCSE) to estimate uncertainty in our quantifications. Since MCSE was in all cases lower than 1\%, we decided to omit it for the sake of brevity.

\section{Results}

\begin{figure}
  \includegraphics[width=\columnwidth]{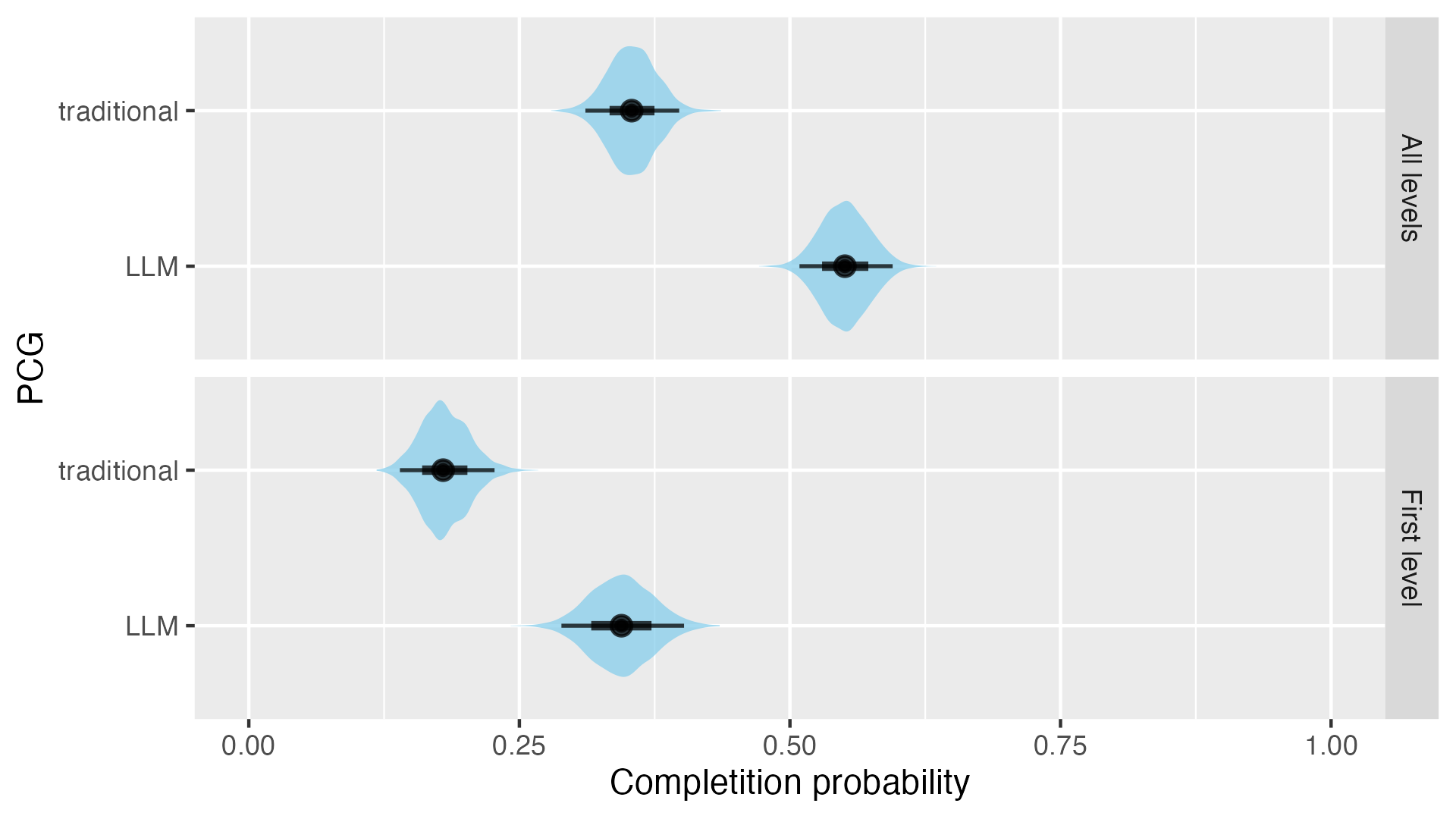} 
\caption{\textbf{Level completion probabilities comparing LLM-generated levels vs. traditional PCG.
} Players were more likely to complete levels generated using LLMs than those generated traditionally. The same holds both when looking at all levels and when looking just at the first level.} \label{completition_rates}
\end{figure}

We compared the results for level completion between the group that was served levels generated with traditional PCG and the group that was served levels generated using LLM PCG. About a third of players ($34\%$) completed the first level that was generated using LLMs, while only $18\%$ completed the traditionally generated first level. Results are displayed in Figure~\ref{completition_rates}. Comparing the data for all levels, the completion rate of levels started for LLMs is $55\%$ and for traditional PCG $35\%$.

The probability that a greater proportion of players would complete any given level when using LLMs, as opposed to the traditionally served level is near certain (P $\approx$ $1.00$). Similarly, the probability that more players would complete the first level when using LLMs compared to the traditionally generated is also extremely high (P $\approx$ $1.00$). 

\begin{figure}
  \includegraphics[width=\columnwidth]{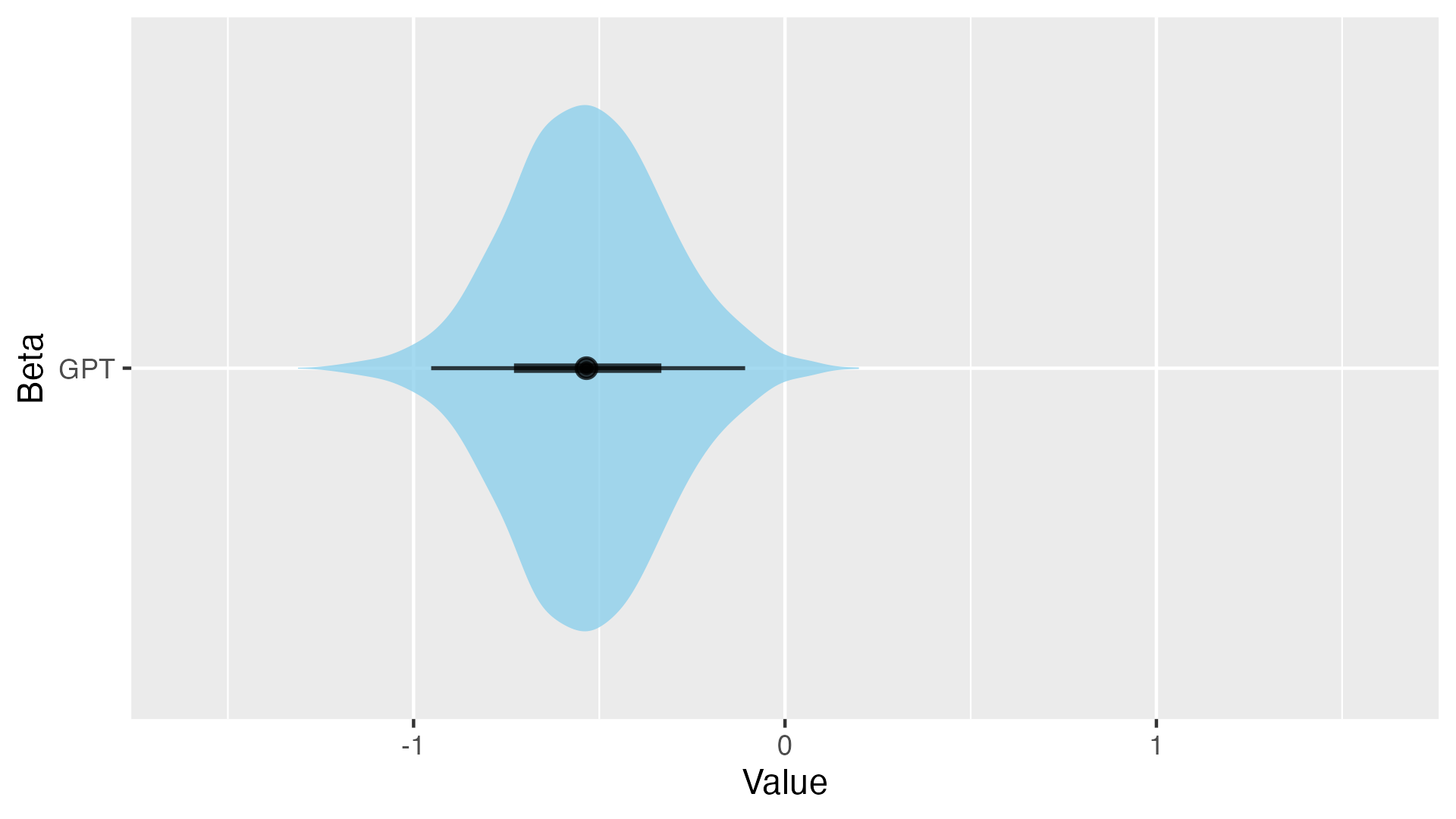} 
\caption{\textbf{A visualization of the beta coefficient, comparing the ratings for LLM-based PCG against traditional PCG, excluding dropouts.} Since the beta coefficient is very likely negative (P $=$ 0.99), we can claim with high confidence that levels generated with traditional PCG have a higher rating than those generated with LLM PCG. Note here that in this analysis we did not account for dropouts -- players who left the game before completing and rating the level.} \label{ratings_beta}
\end{figure}

\begin{figure}
  \includegraphics[width=\columnwidth]{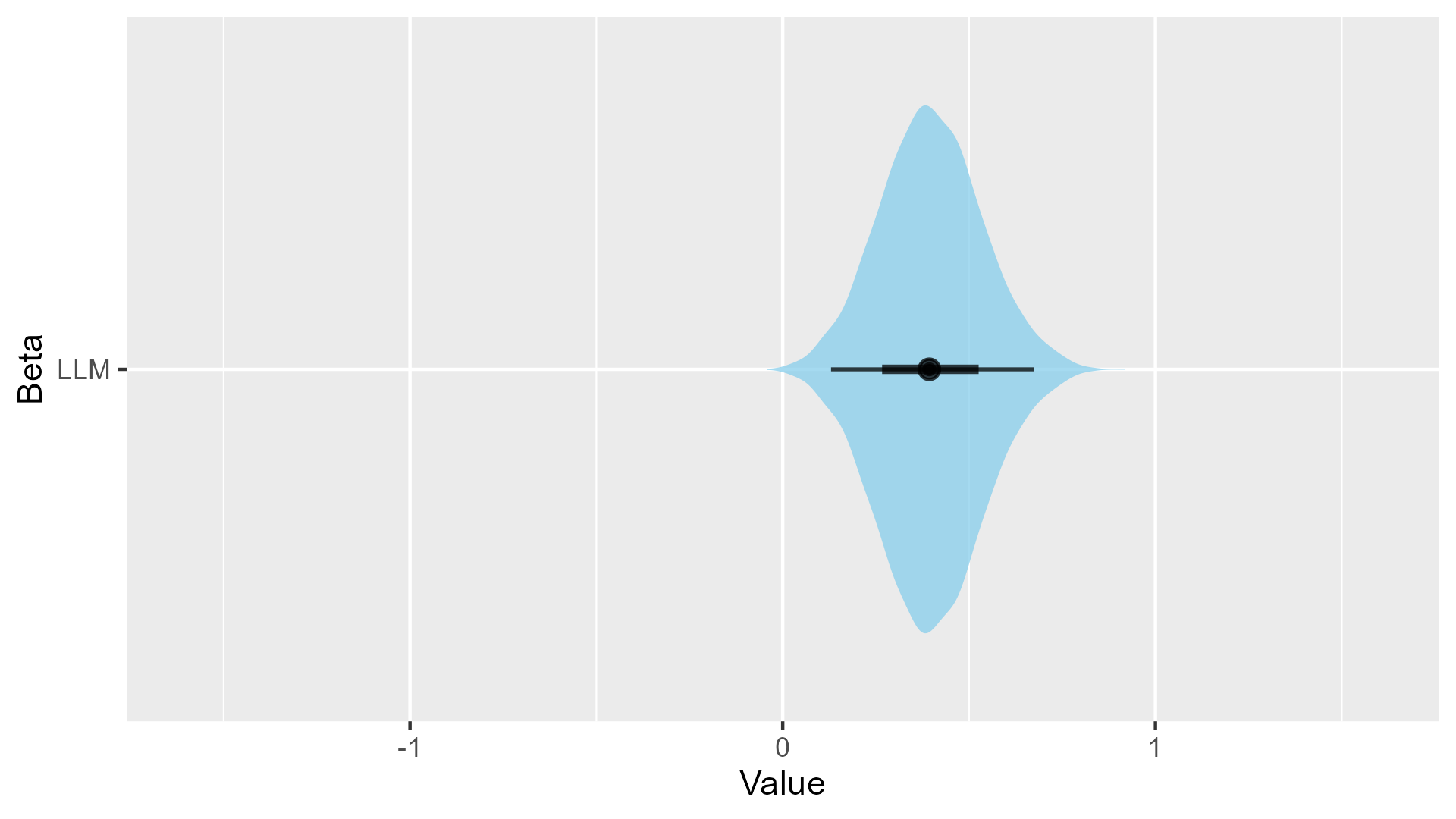} 
\caption{\textbf{A visualization of the beta coefficient, comparing the ratings for LLM-based PCG against traditional PCG, including dropouts.} Since the beta coefficient is very likely positive (P $=$ 0.99), we can claim with high confidence that levels generated with LLM PCG have a higher rating than those generated with traditional PCG when accounting for dropouts. This suggests that a lot of players were very unhappy with traditional PCG and decided to quit the level mid-way.} \label{ratings_beta_abandoned}
\end{figure}

\begin{figure}
  \includegraphics[width=\columnwidth]{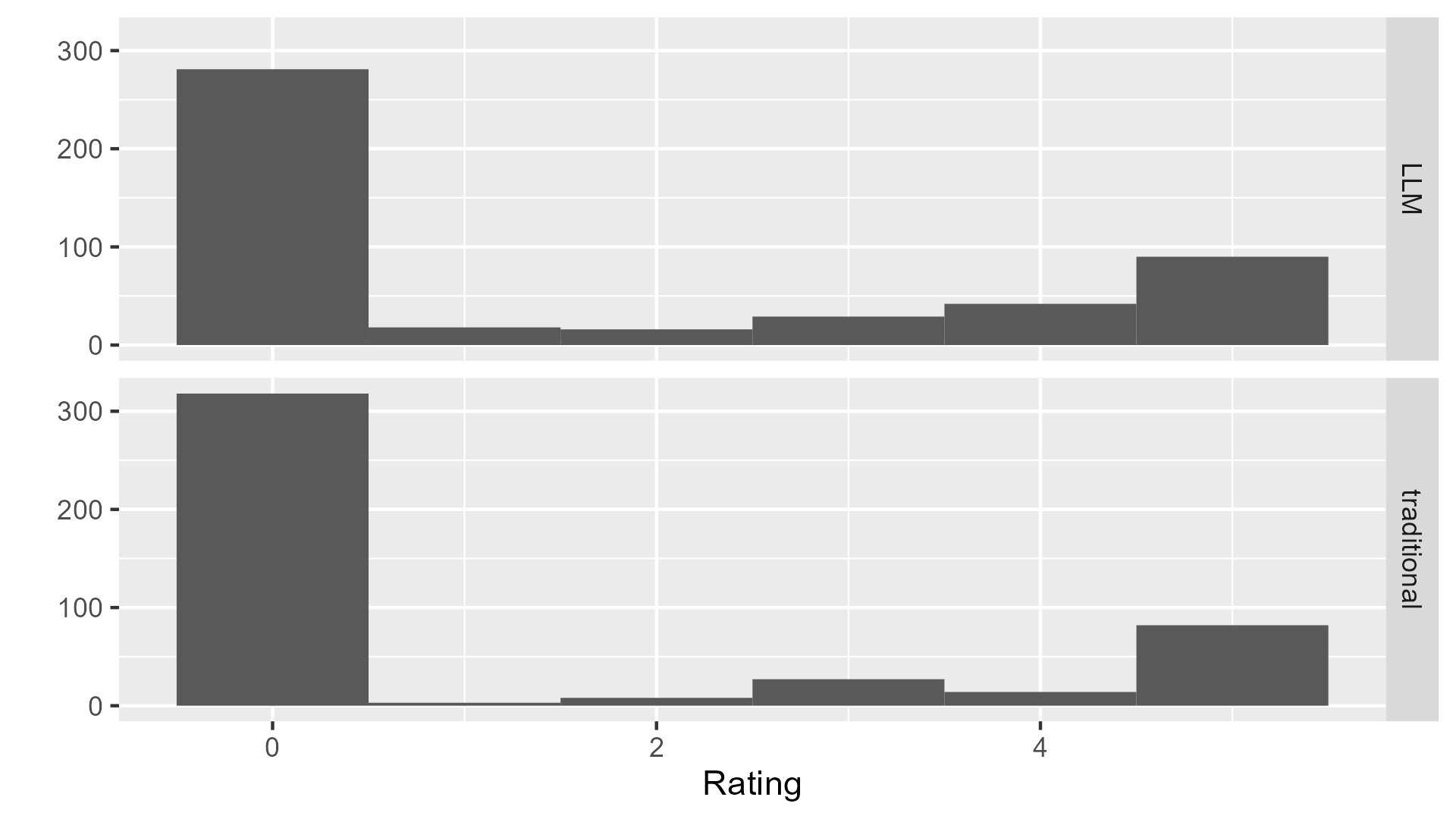} 
\caption{\textbf{This figure indicates that the LLM PCG group had a higher level completion rate, leading to more recorded ratings, including lower ones.} To account for players who exited without completing a level, we assigned a $0$ rating. This adjustment provides a more accurate reflection of player engagement and satisfaction across compared content generation methods.} \label{ratings_histogram}
\end{figure}

Apart from level completion, we also collected data on how much players liked the levels. We asked them to vote on the levels after they completed them. This vote was passed to the LLM and was used to personalize the next levels for the player. Average player ratings were $3.87/5$ for LLM levels and $4.22/5$ for traditional PCG. As shown in Figure~\ref{ratings_beta}, we can say with high confidence (P $=$ $0.99$), that the ratings were lower for LLM-based PCG. As previously shown, the probability that players belonging to the traditional PCG group do not finish a level is higher than for players from the LLM PCG group. One could argue, that these players who are the most disappointed with the level, do not even get to the rating screen but exit the game before that. We call the level runs where this happened dropouts. To account for this, we assigned the rating of 0 to all dropout runs -- runs where a player exited a level before getting to the rating screen. The average rating for LLM-based PCG in this case is $1.59$, and $1.25$ for the traditional PCG. We can claim with high probability (P $=$ 0.99), that LLM PCG generated levels have a higher rating in comparison with traditional PCG generated levels, if we account for dropouts. See Figure~\ref{ratings_beta_abandoned} for a visualization of this analysis. Furthermore, Figure~\ref{ratings_histogram} visualizes the distribution of all ratings.

\begin{figure}
  \includegraphics[width=\columnwidth]{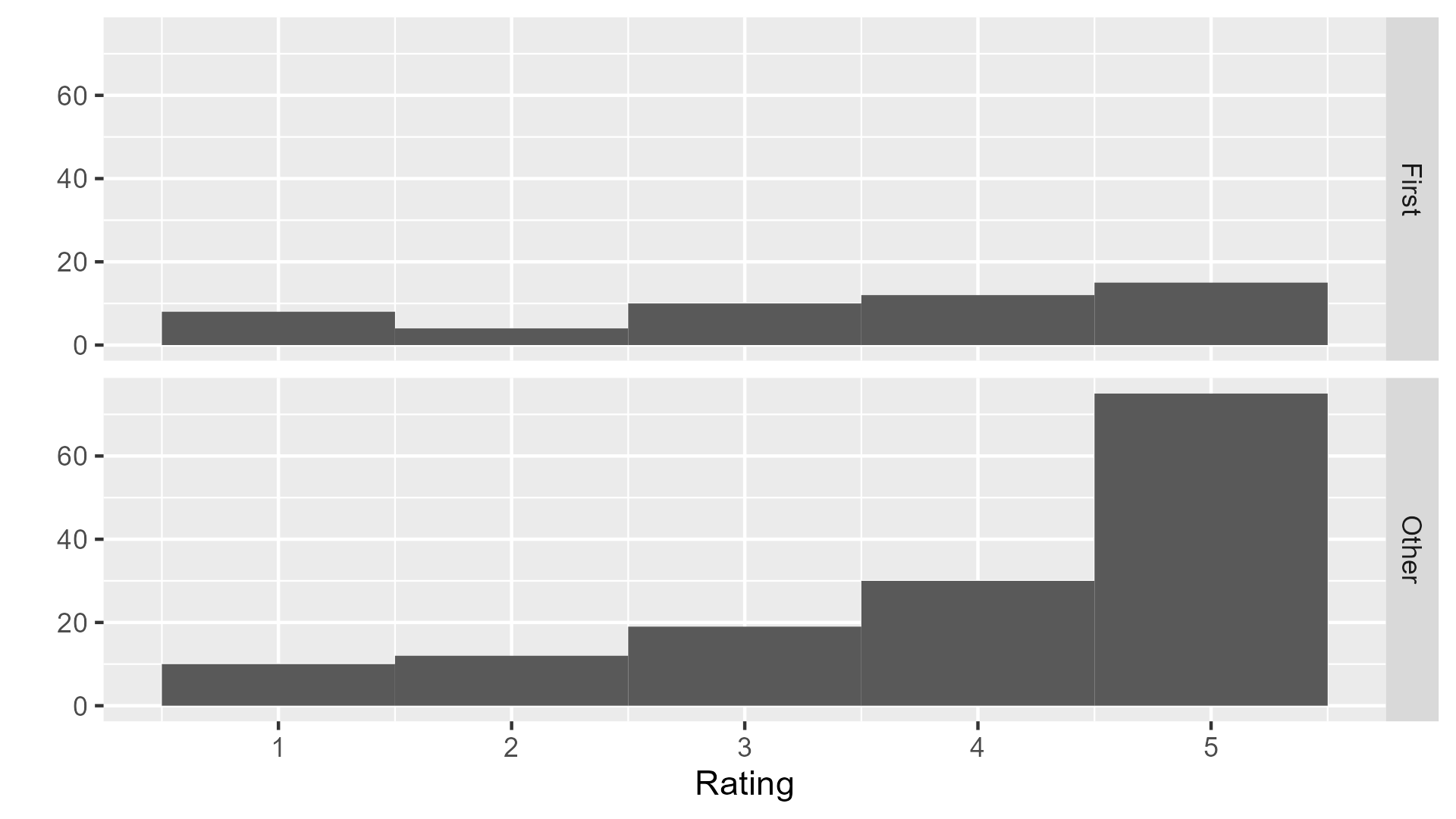}
\caption{\textbf{Comparison of player ratings between the first level and subsequent levels in the LLM PCG group.} Player ratings were higher for levels following the initial one, indicating increased player satisfaction for the LLM-based PCG as the game progressed.\label{ratings_histogram_other_level}}
\end{figure}

As personalization aims to enhance the gaming experience progressively, we also compared the ratings for the initial level against those for subsequent levels. This comparison is depicted in Figure~\ref{ratings_histogram_other_level}. The average rating for the first level stands at $3.45$, while it increases to $4.01$ for subsequent levels. This indicates that the levels after the first are rated significantly higher with an extremely high probability (P $\approx$ 1).

\section{Study limitations}

The weakness of using zero-shot reasoning is that while it offers a simple way to personalize a mobile game, it is limited in how well it can perform compared to custom-trained ML models. To mitigate this shortcoming, our initial multivariate test design included a third group. In this group instructions for the LLM included the comparison of player's data for the completed level against the average, minimum, maximum and median of all the data on this level for players that have previously completed it. This approach would potentially enable our game to leverage collected data to provide increasingly accurate suggestions over time. However, as each level is unique, the approach required clustering to approximate \textit{similar} levels and relying on just the player data until players completed enough levels. Given the added complexities and discretionary decisions we would have to make, we decided to omit an additional group for this experiment. Retrospectively, such a design might be better suited using more objective metrics like level completion, as using data on level ratings, according to our analysis, might not be an accurate reflection of a player's satisfaction with the game.

In our testing, levels generated using traditional PCG received a higher average rating than levels generated using LLM-based PCG. The result surprised us and further analysis has shown a large discrepancy in really low scores for traditional PCG. Rating a level requires level completion and the player's action to cast a vote at the end of the level. As Figure~\ref{ratings_histogram} suggests, once we accounted for the discrepancy by assigning the rating of $0$ to dropouts, we see that the average rating is higher for the LLM-based PCG. As shown in Figure~\ref{completition_rates}, players were more likely to complete levels generated using LLM-based PCG. This suggests that players assigned to the traditional PCG group who did not like the level likely abandoned the game without casting a negative vote. 

We cannot disregard the possibility that dropout rates in the traditional PCG group are higher due to an unsuitable level of difficulty in initial levels. If the levels were too challenging for first-time players, adjusting the difficulty could have reduced the dropout rates and potentially improved the ratings. Additional testing with different difficulty thresholds would be required to completely rule out this possibility. However, even if initial levels in the traditional PCG group were indeed too hard, this again showcases the usefulness of LLM-based PCG as the difficulty problems there seem to be alleviated. As a result game development teams can spend less time on tuning the difficulty of starting levels and spend more resources on other tasks that will make their game better.

We are aware that at the time of testing, using a commercial solution like GPT-4 could incur significant costs, especially when scaling to really large numbers of players. A cheaper solution like GPT-3.5 might produce similar results at lower costs. However, we decided to perform our tests using a state-of-the-art model, as the field is developing rapidly and optimizations and new developments have driven costs down over time in the past.

Our framework encompasses a wide range of settings and parameters, such as which parameters to adjust for the levels, ranges for these parameters, what player types to choose from and how to formulate the instructions for the LLM. We arrived at sensible configurations through a process of experimentation and hands-on gameplay. Our focus was not on the most optimal configuration, but rather on determining whether LLMs can be effectively employed for PCG. While our prototype confirms they can be employed, we are aware there is ample room for further optimizations. 

One area for further optimization could involve experimenting with the predefined list of player types. An alternative approach could involve simplifying the terms or instructing the model to suggest levels based on a traditional difficulty scale, such as \textit{easy, normal,} and \textit{hard}. This modification could potentially offer more straightforward conclusions, although our current methodology has already yielded clear insights.

\section{Discussion}
Our results suggest that LLMs can be used to introduce personalized PCG into level-based games. Furthermore, they can be used without \textit{fine-tuning} by utilising \textit{zero-shot reasoning}, where the model performs prediction based on instructions with zero training examples. The most obvious benefit of using LLMs versus using recommendation algorithms (like matrix factorisation) is that this approach avoids the cold start problem and is able to provide reasonable suggestions based simply on a combination of prompts and gameplay data. This massively reduces the cost and complexity of bringing games with intelligent, ML-based personalized PCG to market while, as our experiment suggests, still offering measurable benefits. Approaches like ours signal a shift to what some are calling \textit{post-training era}, where innovation is driven by applying generative AI to solve practical problems \cite{Gutmans2023RiseGenEng}.

Another interesting finding is the reliability of LLMs when used for PCG. GPT-4's \textit{function calling} capability, which is designed to return data as valid JSON \cite{OpenAIFunctionCalling}, consistently delivers valid levels. There were also no problems when parsing the JSON format, as the output was consistently properly formatted. Our implementation with the Google Play Store-published game and Google Cloud backend was designed to mimic production quality and work without disclaimers. We solved the problem of high latency of the LLM model by generating and serving three levels at once. The game only checked for new levels based on updated predictions after level completion and used a level generated on older data as a fallback, so players never had to wait for the response. For our testing, the cost of using LLMs was low, as the amount of players was limited and prompts were not very long.

\section{Conclusions and Future Work }

We presented a proof-of-concept framework for personalized PCG in mobile games implemented using zero-shot reasoning. The framework is an end-to-end solution for personalized PCG based on LLMs. We validated our framework by developing and publishing a mobile game that interacted with an OpenAI GPT-4 LLM model to provide personalized levels. We used our game to perform a multivariate test, which has shown that players who are served levels created by our framework were significantly more likely to complete a level they started than those served levels generated using traditional PCG. A direct result of our approach is that players are provided with a potentially unlimited, personalized game experience. To address the scarcity of detailed, production-focused research in the area of personalized PCG, our framework is designed to be easily reproducible. 

While there have been developments in real-time content generation using ML, procedurally generated content is typically the same for all players, regardless of their play style. Additionally, PCG using LLMs was not validated in a production environment. A major advantage of using LLMs capable of zero-shot reasoning -- that can generate levels based solely on instructions and do not require additional training -- is that it eliminates the cold-start problem and is able to generate decent personalized levels from the beginning.

Utilising advances in generative AI, particularly zero-shot reasoning capable LLMs like GPT-4, our validated framework will make it easier for game developers to switch from conventional game design to a design that -- by leveraging modern ML approaches -- increases their chance of creating a game that the players will enjoy. Our research aims to bridge the gap between purely academic research, which is often not directly usable by industry and industry research, which has practical goals but is not easily reproducible by developers with limited resources.

Our test has shown that using LLMs in the proposed way is feasible and beneficial, however, it only scratches the surface of what a more complex and more optimized game using our framework can be and achieve. LLMs can be used to personalize any aspect of the game, from difficulty to style and possibly even cross existing genres. We encourage developers to explore our framework to bring forward a type of game that starts the same but is unrecognisable from player to player a few levels in due to continuous personalization.

\section{Data and code availability}
All data and code are available at https://github.com/dhafnar/match3.

\appendices
\section{JSON representation of a level}
\label{appendix:json-representation}

\begin{lstlisting}[style=jsonStyle]
[
  {
    "num_different_pieces": 4,
    "score_goal": 1500,
    "board_width": 6,
    "board_height": 6,
    "num_moves": 30,
    "collection_goals": [
      20,
      25,
      30
    ]
  },
  {
    "num_different_pieces": 5,
    "score_goal": 1800,
    "board_width": 6,
    "board_height": 6,
    "num_moves": 35,
    "collection_goals": [
      25,
      30,
      35
    ]
  },
  {
    "num_different_pieces": 5,
    "score_goal": 2000,
    "board_width": 6,
    "board_height": 6,
    "num_moves": 40,
    "collection_goals": [
      30,
      35,
      40
    ]
  }
]
\end{lstlisting}

\bibliographystyle{IEEEtran}
\bibliography{references}

\begin{thebibliography}{10}
\providecommand{\url}[1]{#1}
\csname url@samestyle\endcsname
\providecommand{\newblock}{\relax}
\providecommand{\bibinfo}[2]{#2}
\providecommand{\BIBentrySTDinterwordspacing}{\spaceskip=0pt\relax}
\providecommand{\BIBentryALTinterwordstretchfactor}{4}
\providecommand{\BIBentryALTinterwordspacing}{\spaceskip=\fontdimen2\font plus
\BIBentryALTinterwordstretchfactor\fontdimen3\font minus \fontdimen4\font\relax}
\providecommand{\BIBforeignlanguage}[2]{{%
\expandafter\ifx\csname l@#1\endcsname\relax
\typeout{** WARNING: IEEEtran.bst: No hyphenation pattern has been}%
\typeout{** loaded for the language `#1'. Using the pattern for}%
\typeout{** the default language instead.}%
\else
\language=\csname l@#1\endcsname
\fi
#2}}
\providecommand{\BIBdecl}{\relax}
\BIBdecl

\bibitem{rajanen2017personalized}
D.~Rajanen and M.~Rajanen, ``Personalized gamification: a model for play data profiling,'' in \emph{Data-driven gamification design: proceedings of the First International Workshop on Data-Driven Gamification Design (DDGD 2017) co-located with 21st International Academic MindTrek Conference (AcademicMindtrek 2017), September 20, Tampere, Finland}.\hskip 1em plus 0.5em minus 0.4em\relax RWTH Aachen University, 2017.

\bibitem{xue2017dynamic}
S.~Xue, M.~Wu, J.~Kolen, N.~Aghdaie, and K.~A. Zaman, ``Dynamic difficulty adjustment for maximized engagement in digital games,'' in \emph{Proceedings of the 26th International Conference on World Wide Web Companion}, 2017, pp. 465--471.

\bibitem{Panteli2023coldstart}
\BIBentryALTinterwordspacing
A.~Panteli and B.~Boutsinas, ``Addressing the cold-start problem in recommender systems based on frequent patterns,'' \emph{Algorithms}, vol.~16, no.~4, 2023. [Online]. Available: \url{https://www.mdpi.com/1999-4893/16/4/182}
\BIBentrySTDinterwordspacing

\bibitem{shin2020playtesting}
Y.~Shin, J.~Kim, K.~Jin, and Y.~B. Kim, ``Playtesting in match 3 game using strategic plays via reinforcement learning,'' \emph{IEEE Access}, vol.~8, pp. 51\,593--51\,600, 2020.

\bibitem{gudmundsson2018human}
S.~F. Gudmundsson, P.~Eisen, E.~Poromaa, A.~Nodet, S.~Purmonen, B.~Kozakowski, R.~Meurling, and L.~Cao, ``Human-like playtesting with deep learning,'' in \emph{2018 IEEE Conference on Computational Intelligence and Games (CIG)}.\hskip 1em plus 0.5em minus 0.4em\relax IEEE, 2018, pp. 1--8.

\bibitem{dale2021gpt}
R.~Dale, ``Gpt-3: What’s it good for?'' \emph{Natural Language Engineering}, vol.~27, no.~1, pp. 113--118, 2021.

\bibitem{vartinen2022generating}
\BIBentryALTinterwordspacing
S.~V{\"a}rtinen, P.~H{\"a}m{\"a}l{\"a}inen, and C.~Guckelsberger, ``Generating role-playing game quests with gpt language models,'' \emph{IEEE Transactions on Games}, 2022. [Online]. Available: \url{https://doi.org/10.17605/OSF.IO/JTQDB}
\BIBentrySTDinterwordspacing

\bibitem{todd2023level}
G.~Todd, S.~Earle, M.~U. Nasir, M.~C. Green, and J.~Togelius, ``Level generation through large language models,'' in \emph{Proceedings of the 18th International Conference on the Foundations of Digital Games}, 2023, pp. 1--8.

\bibitem{Eleti2023FunctionCalling}
A.~Eleti, J.~Harris, and L.~Kilpatrick, ``Function calling and other api updates,'' \url{https://openai.com/blog/function-calling-and-other-api-updates}, March 2023, accessed: 2023-11-28.

\bibitem{Hendrikx2013}
M.~Hendrikx, S.~Meijer, J.~V.~D. Velden, and A.~Iosup, ``Procedural content generation for games: A survey,'' \emph{ACM Transactions on Multimedia Computing, Communications and Applications}, vol.~9, 2 2013.

\bibitem{Perlin1985}
\BIBentryALTinterwordspacing
K.~Perlin, ``An image synthesizer,'' in \emph{Proceedings of the 12th Annual Conference on Computer Graphics and Interactive Techniques}, ser. SIGGRAPH '85.\hskip 1em plus 0.5em minus 0.4em\relax New York, NY, USA: Association for Computing Machinery, 1985, p. 287–296. [Online]. Available: \url{https://doi.org/10.1145/325334.325247}
\BIBentrySTDinterwordspacing

\bibitem{liu2021deep}
J.~Liu, S.~Snodgrass, A.~Khalifa, S.~Risi, G.~N. Yannakakis, and J.~Togelius, ``Deep learning for procedural content generation,'' \emph{Neural Computing and Applications}, vol.~33, no.~1, pp. 19--37, 2021.

\bibitem{Summerville2018}
A.~Summerville, S.~Snodgrass, M.~Guzdial, C.~Holmgård, A.~K. Hoover, A.~Isaksen, A.~Nealen, and J.~Togelius, ``Procedural content generation via machine learning (pcgml),'' \emph{IEEE Transactions on Games}, vol.~10, pp. 257--270, 9 2018.

\bibitem{sudhakaran2023prompt}
S.~Sudhakaran, M.~Gonz{\'a}lez-Duque, C.~Glanois, M.~Freiberger, E.~Najarro, and S.~Risi, ``Prompt-guided level generation,'' in \emph{Proceedings of the Companion Conference on Genetic and Evolutionary Computation}, 2023, pp. 179--182.

\bibitem{togelius2011whatispcg}
\BIBentryALTinterwordspacing
J.~Togelius, E.~Kastbjerg, D.~Schedl, and G.~N. Yannakakis, ``What is procedural content generation? mario on the borderline,'' in \emph{Proceedings of the 2nd International Workshop on Procedural Content Generation in Games}, ser. PCGames '11.\hskip 1em plus 0.5em minus 0.4em\relax New York, NY, USA: Association for Computing Machinery, 2011. [Online]. Available: \url{https://doi.org/10.1145/2000919.2000922}
\BIBentrySTDinterwordspacing

\bibitem{milovsevic2017early}
M.~Milo{\v{s}}evi{\'c}, N.~{\v{Z}}ivi{\'c}, and I.~Andjelkovi{\'c}, ``Early churn prediction with personalized targeting in mobile social games,'' \emph{Expert Systems with Applications}, vol.~83, pp. 326--332, 2017.

\bibitem{radford2019gpt2}
A.~Radford, J.~Wu, R.~Child, D.~Luan, D.~Amodei, and I.~Sutskever, ``Language models are unsupervised multitask learners,'' 2019.

\bibitem{van2021fine}
J.~van Stegeren and J.~My{\'s}liwiec, ``Fine-tuning gpt-2 on annotated rpg quests for npc dialogue generation,'' in \emph{Proceedings of the 16th International Conference on the Foundations of Digital Games}, 2021, pp. 1--8.

\bibitem{taveekitworachai2023chatgpt4pcg}
P.~Taveekitworachai, F.~Abdullah, M.~F. Dewantoro, R.~Thawonmas, J.~Togelius, and J.~Renz, ``Chatgpt4pcg competition: Character-like level generation for science birds,'' 2023.

\bibitem{chernyavskiy2021nlp}
A.~Chernyavskiy, D.~Ilvovsky, and P.~Nakov, ``Transformers: ``the end of history'' for natural language processing?'' in \emph{Machine Learning and Knowledge Discovery in Databases. Research Track}, N.~Oliver, F.~P{\'e}rez-Cruz, S.~Kramer, J.~Read, and J.~A. Lozano, Eds.\hskip 1em plus 0.5em minus 0.4em\relax Cham: Springer International Publishing, 2021, pp. 677--693.

\bibitem{openai2023gpt4}
\BIBentryALTinterwordspacing
OpenAI, ``Gpt-4 technical report,'' \emph{ArXiv}, vol. abs/2303.08774, 2023. [Online]. Available: \url{https://arxiv.org/abs/2303.08774}
\BIBentrySTDinterwordspacing

\bibitem{hamalainen2023llm}
P.~Hämäläinen, M.~Tavast, and A.~Kunnari, ``Evaluating large language models in generating synthetic hci research data: a case study,'' in \emph{Conference on Human Factors in Computing Systems - Proceedings}.\hskip 1em plus 0.5em minus 0.4em\relax Association for Computing Machinery, 4 2023.

\bibitem{radford2018gpt}
A.~Radford, K.~Narasimhan, T.~Salimans, I.~Sutskever \emph{et~al.}, ``Improving language understanding by generative pre-training,'' 2018.

\bibitem{Brown2020gpt3}
\BIBentryALTinterwordspacing
T.~Brown, B.~Mann, N.~Ryder, M.~Subbiah, J.~D. Kaplan, P.~Dhariwal, A.~Neelakantan, P.~Shyam, G.~Sastry, A.~Askell, S.~Agarwal, A.~Herbert-Voss, G.~Krueger, T.~Henighan, R.~Child, A.~Ramesh, D.~Ziegler, J.~Wu, C.~Winter, C.~Hesse, M.~Chen, E.~Sigler, M.~Litwin, S.~Gray, B.~Chess, J.~Clark, C.~Berner, S.~McCandlish, A.~Radford, I.~Sutskever, and D.~Amodei, ``Language models are few-shot learners,'' in \emph{Advances in Neural Information Processing Systems}, H.~Larochelle, M.~Ranzato, R.~Hadsell, M.~Balcan, and H.~Lin, Eds., vol.~33.\hskip 1em plus 0.5em minus 0.4em\relax Curran Associates, Inc., 2020, pp. 1877--1901. [Online]. Available: \url{https://proceedings.neurips.cc/paper_files/paper/2020/file/1457c0d6bfcb4967418bfb8ac142f64a-Paper.pdf}
\BIBentrySTDinterwordspacing

\bibitem{vaswani2017attention}
\BIBentryALTinterwordspacing
A.~Vaswani, N.~Shazeer, N.~Parmar, J.~Uszkoreit, L.~Jones, A.~N. Gomez, L.~Kaiser, and I.~Polosukhin, ``Attention is all you need,'' in \emph{Advances in Neural Information Processing Systems 30}, I.~Guyon, U.~V. Luxburg, S.~Bengio, H.~Wallach, R.~Fergus, S.~Vishwanathan, and R.~Garnett, Eds.\hskip 1em plus 0.5em minus 0.4em\relax Curran Associates, Inc., 2017, p. 5998–6008. [Online]. Available: \url{https://papers.nips.cc/paper/7181-attention-is-all-you-need}
\BIBentrySTDinterwordspacing

\bibitem{lampert2009unseen}
C.~H. Lampert, H.~Nickisch, and S.~Harmeling, ``Learning to detect unseen object classes by between-class attribute transfer,'' in \emph{2009 IEEE Conference on Computer Vision and Pattern Recognition}, 2009, pp. 951--958.

\bibitem{johnson-etal-2017-googles}
\BIBentryALTinterwordspacing
M.~Johnson, M.~Schuster, Q.~V. Le, M.~Krikun, Y.~Wu, Z.~Chen, N.~Thorat, F.~Vi{\'e}gas, M.~Wattenberg, G.~Corrado, M.~Hughes, and J.~Dean, ``{G}oogle{'}s multilingual neural machine translation system: Enabling zero-shot translation,'' \emph{Transactions of the Association for Computational Linguistics}, vol.~5, pp. 339--351, 2017. [Online]. Available: \url{https://aclanthology.org/Q17-1024}
\BIBentrySTDinterwordspacing

\bibitem{lauscher-etal-2020-zero}
\BIBentryALTinterwordspacing
A.~Lauscher, V.~Ravishankar, I.~Vuli{\'c}, and G.~Glava{\v{s}}, ``From zero to hero: {O}n the limitations of zero-shot language transfer with multilingual {T}ransformers,'' in \emph{Proceedings of the 2020 Conference on Empirical Methods in Natural Language Processing (EMNLP)}, B.~Webber, T.~Cohn, Y.~He, and Y.~Liu, Eds.\hskip 1em plus 0.5em minus 0.4em\relax Online: Association for Computational Linguistics, Nov. 2020, pp. 4483--4499. [Online]. Available: \url{https://aclanthology.org/2020.emnlp-main.363}
\BIBentrySTDinterwordspacing

\bibitem{wei2022finetuned}
\BIBentryALTinterwordspacing
J.~Wei, M.~Bosma, V.~Zhao, K.~Guu, A.~W. Yu, B.~Lester, N.~Du, A.~M. Dai, and Q.~V. Le, ``Finetuned language models are zero-shot learners,'' in \emph{International Conference on Learning Representations}, 2022. [Online]. Available: \url{https://openreview.net/forum?id=gEZrGCozdqR}
\BIBentrySTDinterwordspacing

\bibitem{kojima2022zsreason}
\BIBentryALTinterwordspacing
T.~Kojima, S.~S. Gu, M.~Reid, Y.~Matsuo, and Y.~Iwasawa, ``Large language models are zero-shot reasoners,'' in \emph{Advances in Neural Information Processing Systems}, S.~Koyejo, S.~Mohamed, A.~Agarwal, D.~Belgrave, K.~Cho, and A.~Oh, Eds., vol.~35.\hskip 1em plus 0.5em minus 0.4em\relax Curran Associates, Inc., 2022, pp. 22\,199--22\,213. [Online]. Available: \url{https://proceedings.neurips.cc/paper_files/paper/2022/file/8bb0d291acd4acf06ef112099c16f326-Paper-Conference.pdf}
\BIBentrySTDinterwordspacing

\bibitem{wan-etal-2023zsreason}
\BIBentryALTinterwordspacing
X.~Wan, R.~Sun, H.~Dai, S.~Arik, and T.~Pfister, ``Better zero-shot reasoning with self-adaptive prompting,'' in \emph{Findings of the Association for Computational Linguistics: ACL 2023}, A.~Rogers, J.~Boyd-Graber, and N.~Okazaki, Eds.\hskip 1em plus 0.5em minus 0.4em\relax Toronto, Canada: Association for Computational Linguistics, Jul. 2023, pp. 3493--3514. [Online]. Available: \url{https://aclanthology.org/2023.findings-acl.216}
\BIBentrySTDinterwordspacing

\bibitem{haas2014unity}
J.~K. Haas, ``A history of the unity game engine,'' 2014.

\bibitem{lin2020match3}
W.~Lin, ``Make a puzzle match game in unity,'' \url{https://www.udemy.com/course/make-a-puzzle-match-game-in-unity/}, 2020, accessed: 2020-11-15.

\bibitem{OpenAIFunctionCalling}
{OpenAI}, ``Function calling - openai api,'' \url{https://platform.openai.com/docs/guides/function-calling}, 2023, accessed: 2023-12-07.

\bibitem{OpenAI2023TemperatureParameter}
OpenAI, ``How should i set the temperature parameter?'' \url{https://platform.openai.com/docs/guides/text-generation/how-should-i-set-the-temperature-parameter}, 2023, accessed: 2024-03-17.

\bibitem{StanDevelopmentTeam}
\BIBentryALTinterwordspacing
{Stan Development Team}, ``{Stan Modeling Language Users Guide and Reference Manual},'' 2023. [Online]. Available: \url{https://mc-stan.org}
\BIBentrySTDinterwordspacing

\bibitem{bscourse2023}
\BIBentryALTinterwordspacing
J.~Demšar and E.~Štrumbelj, ``Bayesian statistics for computer scientists,'' 2023. [Online]. Available: \url{https://github.com/fri-datascience/course_bs}
\BIBentrySTDinterwordspacing

\bibitem{Gelman2008Weakly}
\BIBentryALTinterwordspacing
A.~Gelman, A.~Jakulin, M.~G. Pittau, and Y.-S. Su, ``{A weakly informative default prior distribution for logistic and other regression models},'' \emph{The Annals of Applied Statistics}, vol.~2, no.~4, pp. 1360 -- 1383, 2008. [Online]. Available: \url{https://doi.org/10.1214/08-AOAS191}
\BIBentrySTDinterwordspacing

\bibitem{Gutmans2023RiseGenEng}
\BIBentryALTinterwordspacing
A.~Gutmans. (2023, June) The rise of geneng: How ai changes the developer role. Accessed: 2023-04-01. [Online]. Available: \url{https://cloud.google.com/blog/products/ai-machine-learning/the-rise-of-geneng-how-ai-changes-the-developer-role}
\BIBentrySTDinterwordspacing

\end{thebibliography}

\vfill

\end{document}